\documentclass[conference,a4paper]{IEEEtran}
\IEEEoverridecommandlockouts

\usepackage{cite}
\usepackage{amsmath,amssymb,amsfonts}
\usepackage{algorithmic}
\usepackage{graphicx}
\usepackage{textcomp}
\usepackage{xcolor}
\usepackage{booktabs}
\usepackage{tabularx}
\usepackage{csquotes}
\usepackage{float}
\usepackage{multirow}
\usepackage{bm}

\def\BibTeX{{\rm B\kern-.05em{\sc i\kern-.025em b}\kern-.08em
    T\kern-.1667em\lower.7ex\hbox{E}\kern-.125emX}}
\begin{document}

\title{Fed-CausalDiff: Decoupled Synchronization for Federated Do-Simulation and Policy Evaluation\\
\thanks{This work was funded by the Akademiaavtalen project ASPIRE.}
}

\author{
\IEEEauthorblockN{pengfei li, Mohammad Khalil}
\IEEEauthorblockA{
\textit{Centre for the Science of Learning \& Technology (SLATE), University of Bergen}\\
Bergen, Norway\\
Email: \{Pengfei.Li, mohammad.khalil\}@uib.no
}
}

\maketitle

\begin{abstract}
While federated learning enables collaborative modelling on decentralised data, standard methods merely fit historical observations. This purely observational approach is fundamentally insufficient for interventional inference and policy evaluation, as sequential actions dynamically alter future states. We propose \textbf{Fed-CausalDiff}, a federated causal diffusion framework for do-simulation. The architecture structurally decomposes the latent state evolution into a global causal score function and a local confounding score function. This design enables \emph{decoupled synchronisation} (DSS), where clients aggregate only the shared causal mechanism while retaining site-specific confounders locally to handle heterogeneity. Experiments on four datasets demonstrate that Fed-CausalDiff achieves better ATE and policy-value estimation accuracy, offering a favorable trade-off between communication cost and inference fidelity.
\end{abstract}


\section{Introduction}

In sectors like multi-center healthcare, algorithmic advertising, and financial risk management, decision systems generate massive sequential logs through \enquote{context-action-feedback} interactions \cite{joachims2018deep}. However, strict data privacy regulations (e.g., GDPR) and ownership barriers often fragment these datasets, preventing cross-domain centralisation \cite{de2020principles}. While Federated Learning enables collaborative modelling \cite{kairouz2021advances}, the industry focus is shifting from simple state prediction to intervention evaluation---specifically, the exploration of counterfactual outcomes under alternative decisions \cite{gao2024causal}. This transition is particularly challenging in long-term sequences, where the tight coupling between actions and states creates complex time-varying confounding. As a result, traditional correlation-based models are prone to systematic bias, which complicates the accurate estimation of intervention effects \cite{shinozaki2020understanding}. Furthermore, because these sequential logs stem from historical policies rather than randomised trials, directly fitting models to observational correlations risks significant off-policy bias when evaluating novel, untested strategies \cite{xie2018off}.

Given this context, existing approaches exhibit distinct limitations. Many federated generative models prioritise synthetic fidelity: FedGAN synchronises generator/discriminator parameters across clients \cite{rasouli2020fedgan}, while TimeGAN combines adversarial loss with stepwise supervision to model time-series distributions \cite{yoon2019time}. However, these models generally aim to fit observational distributions rather than explicitly characterising $do(\cdot)$ intervention semantics, making them ill-suited for reliable closed-loop policy value inference. Conversely, sequential causal inference and OPE offer tools such as doubly robust estimators to assess the returns of new policies on logged data \cite{jiang2016doubly}. Yet, these methods typically lack a sampleable simulator that can progressively generate complete counterfactual trajectories under a specific policy \cite{jiang2016doubly}. While recent federated causal methods (e.g., FedCM) address mechanism identification, they largely neglect sampleable policy simulation \cite{rahman2025fedcm}. Moreover, to circumvent the instability of adversarial training in non-IID settings \cite{amalan2022multi}, we adopt score-based conditional diffusion to achieve robust federated causal dynamic generation.

To bridge these gaps, we propose \textbf{Fed-CausalDiff}, a federated framework for interventional sequence modelling based on conditional latent diffusion.
Fed-CausalDiff models state evolution via conditional denoising (score matching), enabling do-simulation and policy evaluation beyond observational fitting.
To address client heterogeneity, we decompose the score dynamics into a \emph{shared causal} component and a \emph{client-specific confounding} component, and apply partial synchronisation: the server aggregates only the causal score parameters while keeping confounding parameters local.
We train the model with factual pre-training, diffusion-based transition learning, and representation balancing, and evaluate on intervention and offline policy-value metrics against federated generation and counterfactual baselines.

\section{Background}
\subsection{Federated learning and federated generative sequence modelling}
Federated Learning (FL) operates on a paradigm of \enquote{local data retention, on-device training, and server aggregation.} The classic FedAvg balances communication efficiency and scalability by performing local multi-step optimisation and periodically aggregating parameters \cite{mcmahan2017communication}. Amidst non-IID data and system heterogeneity (variations in client compute, participation, or local steps), convergence and robustness remain central challenges. FedProx mitigates training instability arising from statistical and system heterogeneity by incorporating a proximal term into the objective function \cite{li2020federated}. Addressing \enquote{objective inconsistency} caused by disparate local update steps, FedNova rectifies bias and convergence behaviour via normalised aggregation \cite{wang2020tackling}. In FL contexts, generative modelling facilitates \enquote{distribution alignment, data augmentation, cross-institutional sharing,} and \enquote{privacy-preserving synthetic data release.} FedGAN embeds GAN training within federated protocols, enabling collaborative generator-discriminator optimisation across distributed data \cite{rasouli2020fedgan}. In the area of diffusion generative modelling, frameworks such as FedDDPM incorporate the denoising score matching objective of DDPMs into federated protocols \cite{peng2025federated}.

\subsection{Sequential causal inference and counterfactual prediction}
The foundational frameworks of Causal Inference comprise Structural Causal Models (SCMs) and the potential outcomes framework. Pearl systematically codified causal diagrams, structural equations, and counterfactual semantics, offering identification tools such as the do-calculus to bridge observational and interventional distributions \cite{pearl2009causality}. Meanwhile, Hernán and Robins centred on potential outcomes and graphical models, covering inference methods from static to complex longitudinal settings \cite{hernan2010causal}.

Addressing longitudinal data and time-varying treatments, classic approaches such as Marginal Structural Models (MSM) and Inverse Probability of Treatment Weighting (IPTW) construct \enquote{weighted pseudo-randomisation} to estimate causal effects in the presence of time-dependent confounding \cite{robins2000marginal}. Relatedly, the g-formula paradigm emphasises modelling conditional outcome distributions via recursive integration (or simulation) to derive expected counterfactual outcomes under dynamic treatment strategies \cite{hernan2010causal}.

Recently, deep sequence counterfactual prediction methods have emerged for complex high-dimensional longitudinal data. Adopting MSM weighting within a sequence-to-sequence architecture, Recurrent Marginal Structural Networks (RMSN) forecast individual trajectory responses under multi-step treatment plans \cite{lim2018forecasting}. Similarly, Counterfactual Recurrent Networks (CRN) mitigate time-varying confounding bias via adversarially balanced representations to enable time-evolving counterfactual outcome prediction \cite{bica2020estimating}. Closely linked to policy evaluation, OPE estimates the target policy value solely from historical behavioural data when online deployment is unfeasible \cite{uehara2022review}. Representative Doubly Robust (DR) estimators combine importance sampling and direct methods to enhance robustness and reduce variance \cite{jiang2016doubly}. 

\subsection{Federated causal learning and causal discovery}
When data remains distributed across institutions, Federated Causal Inference estimates cross-site causal effects without exchanging individual-level records. Xiong et al. addressed Federated ATE inference amidst heterogeneous populations and treatment mechanisms, proposing a privacy-preserving framework for joint inference \cite{xiong2023federated}. Further work on multi-site observational data established federated inference workflows relying solely on exchanged aggregated statistics \cite{khellaf2025federated}. Beyond federated ATE estimation, recent work has explored richer causal objects and learning paradigms under decentralisation. Khellaf et al. study multi-study federated causal inference beyond simple meta-analysis, comparing one-shot and multi-shot protocols built on plug-in g-formula estimators \cite{khellaf2024federated}.

Federated Causal Discovery (FCD) reconstructs global causal graphs under privacy constraints. Early efforts, such as DS-FCD, integrated differentiable structure learning with federated training to learn causal structures from heterogeneous data without accessing raw local samples \cite{gao2021federated}. In constraint-based approaches, FedC2SL performs skeleton learning and orientation via federated conditional independence testing, enhancing robustness to client disparities under realistic assumptions \cite{wang2023towards}. FedCM learns a proxy SCM via deep generative models and modularises mechanisms between global aggregation and local training \cite{rahman2025fedcm}.

\section{Methodology}

\subsection{Causal Setup and Problem Definition}
Our theoretical framework is grounded in SCMs \cite{pearl2012causal}. By defining endogenous variables through structural equations, we treat the dynamic evolution of subjects as a discrete-time causal system. Given static attributes $X$ and time-varying contexts $A_t$, an intervention policy selects treatment $T_t$, yielding outcome $Y_t$. Central to our approach is the latent state $K_t$, which we structurally decompose into an invariant causal subspace $K^{(c)}_t$ and a site-specific confounding subspace $K^{(s)}_t$. 
The system is governed by two core mechanisms: the \textbf{Probabilistic State Transition} $K_{t+1} \sim p_{\Theta}(K_{t+1} \mid K_t, A_t, \text{do}(T_t))$, modeled via conditional score functions; and the \textbf{Outcome Emission} $Y_t \sim P_{\phi}(Y \mid K_t, A_t, T_t)$.
In the FL setting with $N$ clients, we adopt partial sharing: the server aggregates only the shared parameters $\Theta_c$ associated with the causal score dynamics, while client-specific parameters $\Theta_s^i$ associated with confounding scores are kept local to mitigate heterogeneity \cite{collins2021exploiting}. We correspondingly partition model parameters as $\Theta=\Theta_c \cup \{\Theta_s^i\}_{i=1}^{N}$,
where $\Theta_c$ denotes the \emph{shared sub-network parameters} (including the causal score dynamics and the outcome decoder),
while $\Theta_s^i$ captures client-specific confounding scores.

\subsection{The Fed-CausalDiff Framework}
\begin{figure*}[t]
    \centering
    \includegraphics[width=\linewidth]{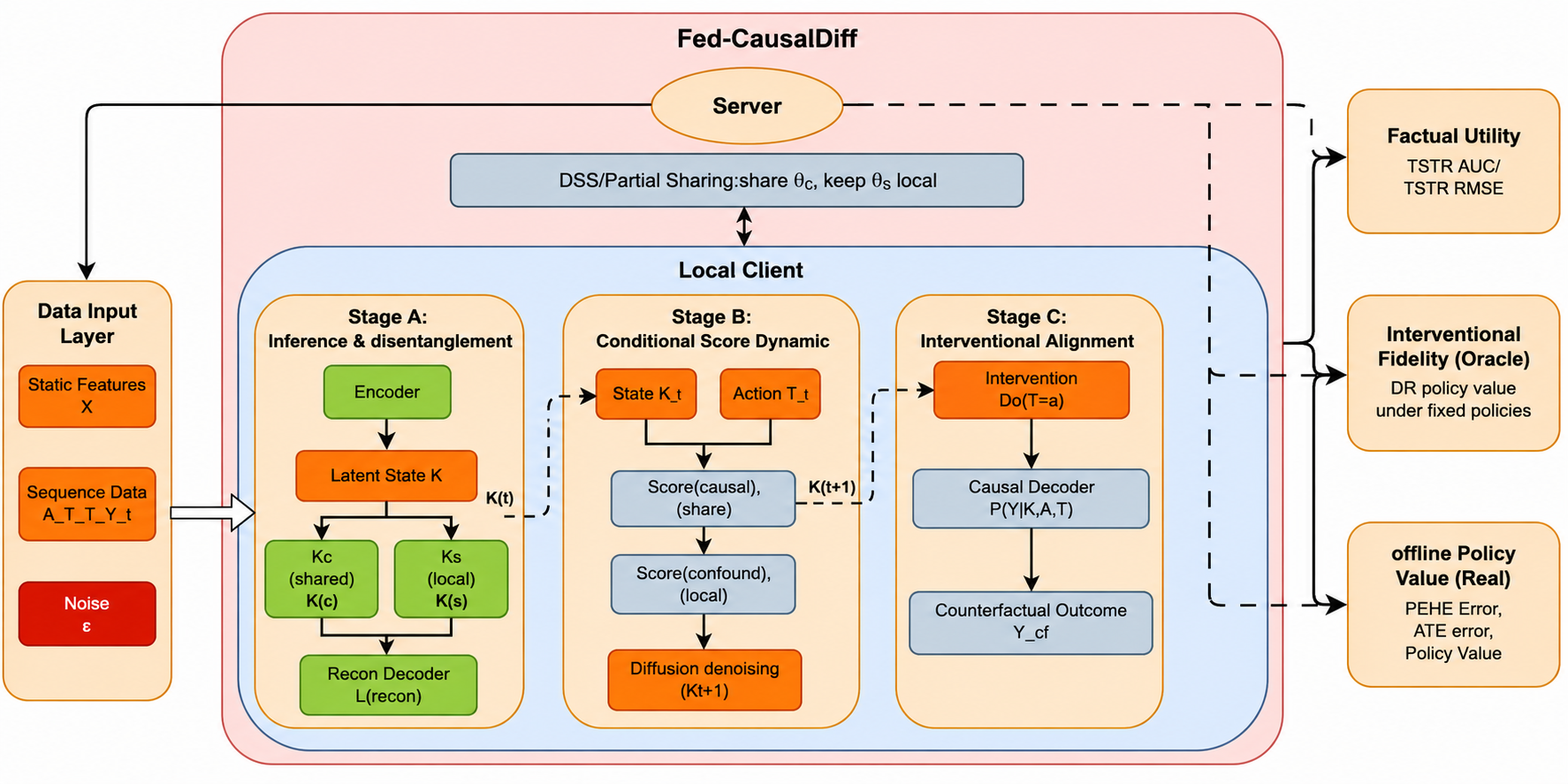}
    \caption{Structure of the Fed-CausalDiff design.}
    \label{fig:exp-setup1}
\end{figure*}

To stabilise the training of complex causal mechanisms within a federated environment, Fed-CausalDiff employs a three-stage decoupled training strategy. 
Crucially, we structurally partition the latent space into a causal subspace $K^{(c)}$ (globally shared) and a confounding subspace $K^{(s)}$ (locally retained). At each communication round, clients update $(\Theta_c,\Theta_s^i)$ locally but only transmit $\Theta_c$ to the server. The server aggregates $\Theta_c$ (FedAvg) and broadcasts the updated $\Theta_c$ back, while $\Theta_s^i$ remains local. The overall three-stage training pipeline is summarised in Fig.~\ref{fig:exp-setup1}.
We train the model locally in three phases (inference, diffusion dynamics, and decoding), and apply DSS at each communication round by synchronising only the shared parameters $\Theta_c$.

\textbf{Phase A: Latent Inference \& Disentanglement.} 
This phase maps observed trajectories to the partitioned latent space $K = [K^{(c)}, K^{(s)}]$ using a recurrent encoder. By minimising trajectory reconstruction error, we ensure $K$ effectively preserves historical context while establishing the initial embedding for both invariant causal factors and site-specific confounders.

\textbf{Phase B: Conditional Score Dynamics.}
Instead of deterministic transitions, we model state evolution as a conditional denoising process. The transition from $K_t$ to $K_{t+1}$ is governed by a score function decomposed into global and local terms:
\begin{equation}
\nabla_{K} \log p(K_{t+1}|K_t) \approx S_{\text{causal}}(\cdot; \Theta_c) + S_{\text{confound}}(\cdot; \Theta_s^i)
\end{equation}
Here, $S_{\text{causal}}$ captures shared causal laws conditioned on interventions $T_t$, while $S_{\text{confound}}$ accounts for local heterogeneity. This decomposition is intended to separate shared causal dynamics from site-specific nuisance, facilitating robust interventional rollouts under client heterogeneity.

\textbf{Phase C: Causal Decoder \& Decoupled Aggregation.} 
This stage employs a decoder $P_{\phi}$ to translate latent states into observable outcomes. 
We correspondingly partition model parameters as $\Theta=\Theta_c \cup \{\Theta_s^i\}_{i=1}^{N}$,
where $\Theta_c$ denotes the \emph{shared sub-network parameters} (including the causal score dynamics and the outcome decoder),
while $\Theta_s^i$ captures client-specific confounding scores.

The three phases play complementary roles. Phase A establishes a disentangled latent space where $K^{(c)}$ captures invariant drivers while $K^{(s)}$ absorbs client-specific heterogeneity, improving cross-client transfer. Phase B learns intervention-conditioned score dynamics, enabling stable stochastic rollouts and more faithful $do(\cdot)$ responses under non-IID settings. Phase C calibrates the emission decoder and enforces decoupled aggregation, so that global updates improve shared causal dynamics without overwriting local confounding factors.

\subsection{Client-wise objectives \& Implicit Global Consistency}
To balance the fidelity of observational data with causal consistency, the local training objective $\mathcal{L}$ replaces unstable adversarial losses with robust score matching constraints:
$$
\mathcal{L} = \underbrace{\mathcal{L}_{recon} + \lambda_{diff} \mathcal{L}_{diff}}_{\text{Generative Fidelity}} + \underbrace{\lambda_1 \mathcal{L}_{align} + \lambda_2 \mathcal{L}_{dis} + \lambda_3 \mathcal{L}_{cf}}_{\text{Causal Regularization}}
$$

\noindent\textbf{Fidelity Objectives:} 
We employ Reconstruction Loss ($\mathcal{L}_{recon}$) to anchor the encoder/decoder. The core dynamics are trained via Diffusion Denoising Loss ($\mathcal{L}_{diff}$):
$$
\mathcal{L}_{diff} = \mathbb{E}_{t,n,\epsilon}[\|\epsilon - (\epsilon_{\theta_c}(K_{t+1}^{(n)}, \dots) + \epsilon_{\theta_s^i}(K_{t+1}^{(n)}, \dots))\|_2^2]
$$
This minimises the error between the added noise $\epsilon$ and the joint prediction from global and local score networks.

\noindent\textbf{Causal Regularization:} 
To mitigate selection bias, we minimise the MMD distance between factual and interventional distributions via \textit{Intervention Alignment} ($\mathcal{L}_{align}$). Crucially, to enforce our decoupled strategy, we implement a \textit{Disentanglement} objective ($\mathcal{L}_{dis}$): a Gradient Reversal Layer (GRL) discourages the encoder's $K^{(c)}$ from encoding \emph{site-specific} artifacts, while a parallel head encourages $K^{(s)}$ to capture nuisance factors.
Finally, \textit{Counterfactual Consistency} ($\mathcal{L}_{cf}$) penalises deviations in historical representations under varying future interventions.

\noindent\textbf{Server-wise Consistency:}
Clients optimise local objectives on their private data, while the server enforces global consistency through DSS by aggregating only the shared parameters $\Theta_c$ and keeping $\Theta_s^i$ local.

\section{Experiment Setup}
\subsection{Datasets}
Our evaluation spans four datasets, encompassing both semi-synthetic and real-world sequential logs:
\begin{itemize}
\item \textbf{DKT-Synth (Synthetic)}: Derived from a pre-trained LSTM-DKT oracle \cite{ghosh2020context}, this semi-synthetic dataset provides ground-truth counterfactual outcomes ($Y_{cf}$), enabling precise causal evaluation.
It consists of student-item interaction sequences in which actions correspond to practice opportunities, and outcomes reflect the oracle-generated mastery response under the interventions.

\item \textbf{Statics2011}: A collection of step-level interaction data from a university engineering statics course \cite{ghosh2020context}.
Each sequence records a learner's problem-solving trajectory over time, capturing temporally ordered contexts, actions, and performance signals suitable for sequential intervention modelling.

\item \textbf{Diabetes-130 (1999--2008)}: A public dataset comprising ten years of clinical hospitalisation records across 130 U.S. hospitals \cite{strack2014impact}.

\item \textbf{Open Bandit Dataset}: Real-world logged bandit data from ZOZOTOWN \cite{saito2020open}.
It contains impression-level logs with stochastic action selection under a historical logging policy, together with observed rewards, making it a standard benchmark for offline policy evaluation under distribution shift.
\end{itemize}
\begin{table}[t]
\caption{Datasets after preprocessing into the unified $(X,A,T,Y)$.}
\label{tab:datasets}
\centering
\small
\setlength{\tabcolsep}{4pt}
\begin{tabular}{lccccc}
\hline
Dataset & \#Seqs & $L_{\max}$ & $d_x$ & $d_a$ & $|T|$ \\
\hline
DKT-Synth & 20000 & 50 & 4 & 1 & 50 \\
Statics2011 & 1134 & 200 & 6 & 1 & 1224 \\
Diabetes-130 & 71523 & 20 & 11 & 74 & 145 \\
OpenBandit & 10000 & 50 & 4 & 18 & 8089 \\
\hline
\end{tabular}
\end{table}

\paragraph{Reference policy family for OPE.}
Table~I reports the full cardinality $|T|$ of the discretised action vocabulary used by the generative model. For offline policy evaluation in Fig.~2, however, we do not enumerate the full multi-action space, which would be computationally expensive and, on several datasets, poorly supported by logged propensities. Instead, for each dataset, we restrict evaluation to a binary anchor-action subset with adequate empirical support. Concretely, among the valid discretised actions, we select two anchor actions with the strongest logged support (and thus the most reliable overlap for DR-based evaluation), denoted by $a_{\mathrm{off}}$ and $a_{\mathrm{on}}$. The labels ``off'' and ``on'' are nominal and do not imply an ordinal treatment intensity; they simply index the fixed binary action subset used to instantiate the reference policy family. We then define four target policies: Never ($T_t \equiv a_{\mathrm{off}}$), Always ($T_t \equiv a_{\mathrm{on}}$), Early-on, and Late-on. Therefore, Fig.~2 should be interpreted as a controlled cross-dataset benchmark on a support-aware binary action subset, rather than an exhaustive evaluation over all $|T|$ possible actions.

\paragraph{Feature engineering and sequence construction.}
For each dataset, raw logs are transformed into the unified schema $(X, A_{1:L}, T_{1:L}, Y_{1:L}, M_{1:L})$ via a common preprocessing pipeline.
Trajectories are first grouped by an entity identifier (e.g., student/user/patient), ordered chronologically, and then truncated or padded to $L_{\max}$, with $M_t\in\{0,1\}$ indicating valid steps.
We discretise $T_t$ into a categorical action ID to support $do(T_t=a)$ interventions, while continuous covariates are normalised (z-score) using training-split statistics.
Categorical covariates are encoded via one-hot (small cardinality) or integer IDs with embeddings (large cardinality), and missing values are handled by explicit ``missing'' indicators when appropriate.

Across datasets, $X$ collects time-invariant attributes (e.g., demographics or baseline profiles), whereas $A_t$ captures the time-varying context available at decision time.
In DKT-Synth and Statics2011, $A_t$ contains the item/step identifier (and optional concept tags), $T_t$ represents the intervenable decision signal (e.g., assigned difficulty/condition), and $Y_t$ corresponds to step correctness.
For Diabetes-130, we construct patient-level sequences by ordering encounters and aggregating encounter attributes: $X$ includes static demographics (e.g., age group, gender, race), $A_t$ aggregates encounter-level clinical context (e.g., diagnoses, labs and medication indicators), and $T_t$ corresponds to discretised treatment/action states derived from medication and management variables; $Y_t$ is defined as the target outcome (e.g., readmission indicator) at each encounter.
For OpenBandit, we treat each user's interaction history as a trajectory; $A_t$ is the provided context feature vector, $T_t$ is the displayed item/action ID, and $Y_t$ is the observed feedback (click reward).
When available (e.g., OpenBandit), logged propensities are retained for DR-based policy evaluation but are not transmitted across clients.

Finally, after sequence construction, we simulate federated heterogeneity by partitioning the preprocessed trajectories into $N=10$ clients using a common Dirichlet non-IID split with concentration parameter $\alpha=10$. This partition rule is applied uniformly across all datasets and all compared methods. The resulting client allocation introduces controlled statistical heterogeneity under a unified federated protocol, while dataset-specific semantics are preserved by the preprocessing definitions of $T_t$ and $Y_t$ described above.

\subsection{Baselines}
We benchmark Fed-CausalDiff against the following four baseline models:
\begin{itemize}
\item \textbf{RCGAN:} Recurrent Conditional GAN (RCGAN) \cite{esteban2017real} serves as our sequence generation baseline. It trains an RNN-based generator and discriminator in a conditional setup to match the distribution of the observational trajectory. As it is not designed for interventional semantics, it provides a strong reference for factual realism but may be insufficient for $do(\cdot)$-level fidelity.

\item \textbf{TimeGAN:} A classic benchmark for time-series generation, TimeGAN \cite{yoon2019time} preserves temporal dynamics by jointly optimising supervised and adversarial losses within a learned embedding space. Its supervised component encourages realistic step-to-step transitions, while adversarial training aligns the generated and real sequence distributions, making it a widely adopted baseline for downstream utility under the TSTR protocol.

\item \textbf{CRN:} CRN \cite{bica2020estimating} is a representative method for sequential counterfactual inference. It learns balanced representations via adversarial training to mitigate time-varying confounding and predicts potential outcomes under alternative treatment sequences. Unlike generative baselines, CRN is effect-estimation oriented and thus provides a strong reference for counterfactual prediction quality.

\item \textbf{FedCM:} This approach approximates underlying SCMs by training deep causal generative models in a federated setting \cite{rahman2025fedcm}. It explicitly targets causal mechanism learning under decentralised data and thus constitutes a close federated causal baseline. 
\end{itemize}
The specific parameters for the federated experiment are shown in ~\ref{tab:fed_protocol}.

\begin{table}[t]
\centering
\small
\caption{Federated training protocol used for all compared methods.}
\label{tab:fed_protocol}
\setlength{\tabcolsep}{4pt} 
\begin{tabularx}{\columnwidth}{@{}l X@{}} 
\toprule
Protocol item & Setting \\
\midrule
Clients & 10 \\
Partition rule & Dirichlet non-IID split ($\alpha=10$) \\
Participation rate & $5/10=0.5$ clients per round \\
Local epochs & 1 \\
Aggregation & Sample-size weighted FedAvg \\
Communication accounting & Pplink/downlink $=$ selected clients $\times$ synchronized parameter size\\
\bottomrule
\end{tabularx}
\end{table}

\subsection{Metrics}
We evaluate factual fidelity using \textbf{Factual AUC} \cite{piech2015deep} (the predictive discrimination of factual outcomes under observed trajectories), and the \textbf{TSTR} protocol \cite{esteban2017real}, where a downstream predictor is trained on synthetic sequences and tested on real data to quantify utility preservation. For causal utility on semi-synthetic benchmarks with counterfactual ground truth, we report \textbf{PEHE} \cite{shalit2017estimating}, which measures individual-level heterogeneous effect error, and \textbf{ATE absolute error}, which quantifies deviation in the population average treatment effect. For real-world scenarios without counterfactual labels, we employ \textbf{offline policy evaluation (OPE)} and estimate policy returns via the \textbf{doubly robust (DR)} estimator \cite{jiang2016doubly}, which combines propensity-based reweighting with a direct outcome model to reduce bias, and we further report \textbf{policy value absolute error} when a reference value is available (e.g., oracle or semi-synthetic), measuring the discrepancy between simulated and reference policy returns.

\section{Results}
\subsection{Interventional Fidelity on Oracle}

\begin{table}[t]
\centering
\small
\setlength{\tabcolsep}{3.2pt}
\caption{Interventional fidelity on \textsc{DKT-Synth} with counterfactual ground truth (lower is better). \textbf{w/o DSS} uses full synchronization of all trainable parameters; \textbf{w/o $L_{\mathrm{dis}}$} disables the disentanglement loss. Best results are \textbf{bolded}.}
\label{tab:results_interventional_oracle}
\begin{tabularx}{\linewidth}{lccc}
\toprule
Model & PEHE$\downarrow$ & ATE$\downarrow$ & Value$\downarrow$ \\
\midrule
Fed-CausalDiff  & \textbf{0.057$\pm$0.01} & \textbf{0.045$\pm$0.00} & \textbf{0.049$\pm$0.01} \\
\quad w/o DSS (full sync) & 0.065 & 0.088 & 0.116 \\
\quad w/o $L_{\mathrm{dis}}$ & 0.064 & 0.086 & 0.117 \\
\midrule
FedCM & 0.101$\pm$0.02 & 0.129$\pm$0.01 & 0.099$\pm$0.01 \\
CRN & 0.082$\pm$0.00 & 0.133$\pm$0.02 & 0.073$\pm$0.01 \\
TimeGAN & 0.205$\pm$0.01 & \textbf{0.045$\pm$0.00} & 0.247$\pm$0.03 \\
RCGAN & 0.124$\pm$0.02 & 0.094$\pm$0.01 & 0.115$\pm$0.02 \\
\bottomrule
\end{tabularx}
\end{table}

This section evaluates intervention fidelity on the DKT-Synth semi-synthetic dataset, where a pre-trained DKT oracle provides ground-truth counterfactuals, enabling a direct comparison of causal effect and policy value recovery across methods. As shown in Table \ref{tab:results_interventional_oracle}, we measure performance using three metrics: PEHE, ATE absolute error, and Policy value absolute error, with lower values indicating more accurate modelling of $do(\cdot)$ responses. Overall, Fed-CausalDiff achieves the tied-best performance across all three metrics, including PEHE (0.057$\pm$0.01), ATE (0.045$\pm$0.00), and policy value error (0.049$\pm$0.01), demonstrating that its aggregated global model more stably aligns intervention responses with long-term returns. In contrast, while TimeGAN performs well on ATE, its significantly higher PEHE and policy-value errors suggest that merely fitting the distributions of the observed sequences is insufficient to guarantee consistent intervention semantics and closed-loop value. Similarly, although CRN shows competitive PEHE, its higher ATE and value errors indicate that methods oriented towards effect estimation do not necessarily translate to value consistency in a sampleable simulator. Moreover, both ablations consistently worsen intervention fidelity: \textbf{w/o DSS (full sync)} increases PEHE (0.057$\rightarrow$0.065) and substantially enlarges ATE/value errors (0.045$\rightarrow$0.088, 0.049$\rightarrow$0.116). 
Disabling disentanglement (\textbf{w/o $L_{\mathrm{dis}}$}) shows a similar trend (PEHE 0.064, ATE 0.086, value error 0.117), highlighting the contribution of DSS and $L_{\mathrm{dis}}$ to stable causal and value recovery.

\subsection{Offline Policy Value Difference}
\begin{figure*}[t]
    \centering
    \includegraphics[width=0.7\linewidth]{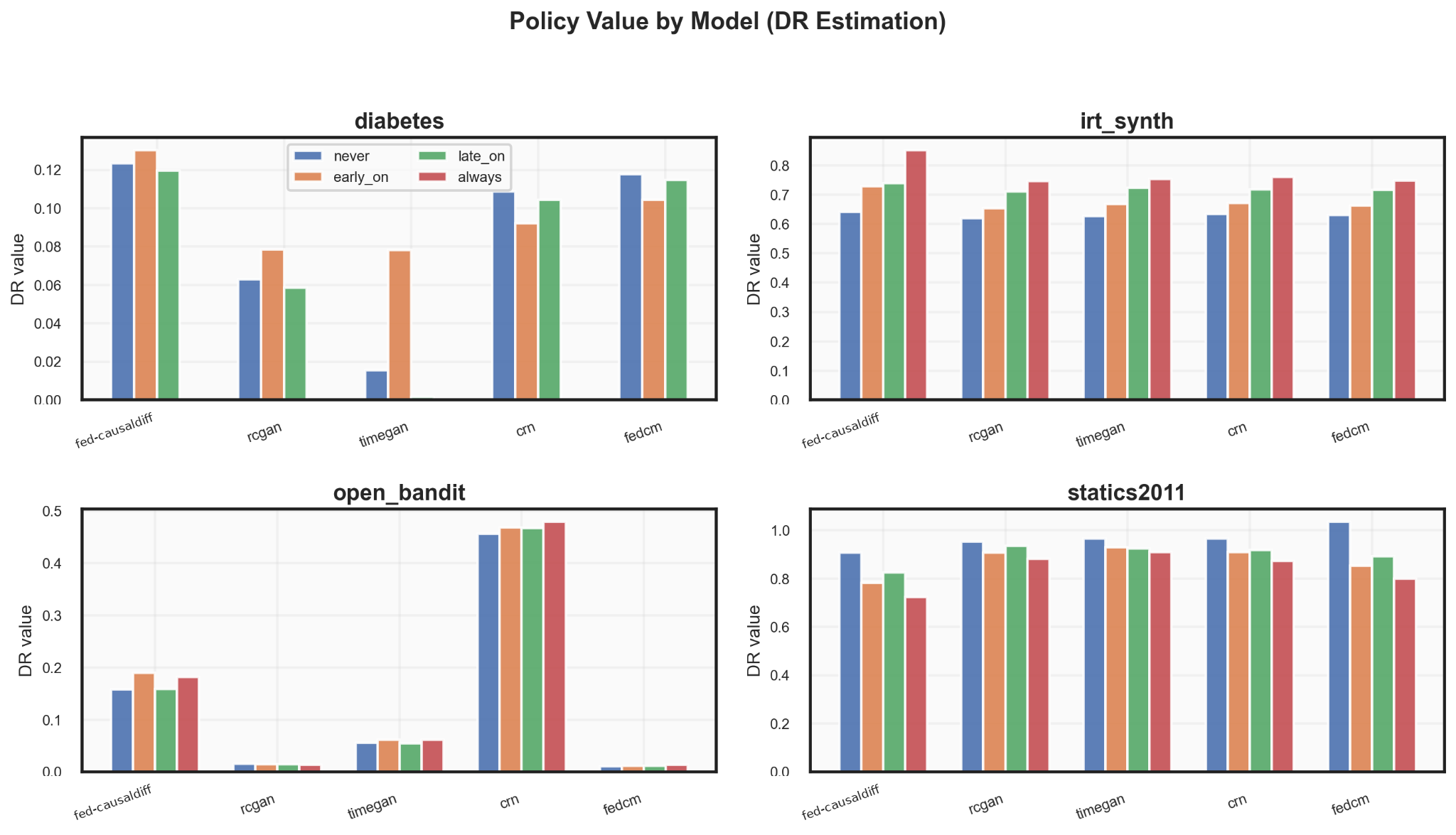}
    \caption{DR offline policy value estimates under four fixed policies (Never, Always, Early-on, Late-on) across the four datasets. Higher indicates better estimated policy performance.}
    \label{fig:dr-policy-values}
\end{figure*}
As illustrated in Figure \ref{fig:dr-policy-values}, we employ DR offline evaluation to estimate the expected returns of four fixed target policies across four datasets, where higher values indicate superior estimated policy performance. On the Diabetes dataset, Fed-CausalDiff consistently yields higher value estimates across the three evaluatable policies---Never (0.123), Early-on (0.130), and Late-on (0.120)---achieving a mean of 0.126. It outperforms both FedCM/CRN and generative baselines, suggesting a superior ability to capture the long-term causal impact of actions on outcomes (note that the \enquote{Always} policy resulted in NaN values for the DR estimator due to insufficient data support). Similarly, on DKT-Synth, Fed-CausalDiff attains the highest mean return (0.740) and demonstrates clear differentiation between policies, peaking at 0.853 under the \enquote{Always} condition. In contrast, the OpenBandit dataset presents a logged bandit scenario characterized by a high-dimensional action space (large $|T|$) and strong dependence on propensity scores. Here, CRN achieves significantly higher DR values (approximately 0.46--0.48) than all other models, while Fed-CausalDiff performs at a moderate level (mean 0.172), slightly surpassing the remaining baselines. Finally, on Statics2011, Fed-CausalDiff yields generally lower estimates (mean 0.809) compared to TimeGAN and CRN. This discrepancy may be attributed to estimation variance and regularisation bias stemming from the dataset's smaller sample size and complex action space.

\subsection{Downstream Utility}

\begin{table}[t]
\centering
\small
\caption{Factual fidelity and TSTR results on current runs. Each entry reports mean$\pm$std across seeds. Best results per dataset and metric are \textbf{bolded}.}
\label{tab:results_factual_trts_tstr}
\setlength{\tabcolsep}{5pt}
\begin{tabular}{llcc}
\toprule
Dataset & Model & TSTR-AUC$\uparrow$ & TSTR-RMSE$\downarrow$ \\
\midrule
Diabetes & Fed-CausalDiff & 0.565$\pm$0.010 & 0.498$\pm$0.059 \\
 & CRN & 0.521$\pm$0.011 & 0.569$\pm$0.013 \\
 & TimeGAN & 0.557$\pm$0.017 & 0.492$\pm$0.027 \\
 & RCGAN & \textbf{0.577$\pm$0.009} & \textbf{0.479$\pm$0.020} \\
 & FedCM & 0.570$\pm$0.005 & 0.552$\pm$0.043 \\
\midrule
DKT-Synth & Fed-CausalDiff & 0.617$\pm$0.007 & 0.509$\pm$0.013 \\
 & CRN & 0.530$\pm$0.012 & 0.504$\pm$0.003 \\
 & TimeGAN & 0.584$\pm$0.054 & 0.513$\pm$0.022 \\
 & RCGAN & \textbf{0.627$\pm$0.008} & 0.492$\pm$0.003 \\
 & FedCM & 0.623$\pm$0.003 & \textbf{0.490$\pm$0.005} \\
\midrule
OpenBandit & Fed-CausalDiff & 0.505$\pm$0.004 & \textbf{0.504$\pm$0.021} \\
 & CRN & 0.497$\pm$0.020 & 0.525$\pm$0.012 \\
 & TimeGAN & 0.510$\pm$0.018 & 0.538$\pm$0.026 \\
 & RCGAN & 0.495$\pm$0.015 & 0.557$\pm$0.040 \\
 & FedCM & \textbf{0.535$\pm$0.021} & 0.525$\pm$0.097 \\
\midrule
Statics2011 & Fed-CausalDiff & 0.562$\pm$0.025 & \textbf{0.390$\pm$0.011} \\
 & CRN & 0.623$\pm$0.01 & 0.471$\pm$0.03 \\
 & TimeGAN & \textbf{0.623$\pm$0.009} & 0.471$\pm$0.03 \\
 & RCGAN & 0.579$\pm$0.036 & 0.422$\pm$0.019 \\
 & FedCM & 0.56$\pm$0.008 & 0.471$\pm$0.3 \\
\bottomrule
\end{tabular}
\end{table}

Table \ref{tab:results_factual_trts_tstr} corroborates that Fed-CausalDiff maintains downstream utility comparable to leading generative and causal baselines, demonstrating that the integration of interventional constraints does not significantly compromise factual fidelity. While generative models like RCGAN and FedCM naturally exhibit an inherent advantage in purely factual metrics, securing top TSTR-AUC and RMSE scores on Diabetes and DKT-Synth, Fed-CausalDiff retains near-optimal performance (e.g., Diabetes AUC 0.565, DKT-Synth AUC 0.617), thereby establishing a robust foundation for causal evaluation. Conversely, on the OpenBandit dataset, Fed-CausalDiff achieves the lowest TSTR-RMSE (0.504), indicating superior predictive stability despite the high class imbalance that hinders all models. Notably, on Statics2011, Fed-CausalDiff significantly outperforms other methods with a TSTR-RMSE of 0.390; although its AUC is marginally lower than the top performer, this underscores its superior error control in scenarios with small sample sizes and strong sequential dependencies.

\subsection{System Efficiency: Convergence and Communication--Performance Trade-off}
\label{sec:sys_efficiency}

\paragraph{Rounds-to-target interventional accuracy.}
\begin{figure}[t]
    \centering
    \includegraphics[width=\linewidth]{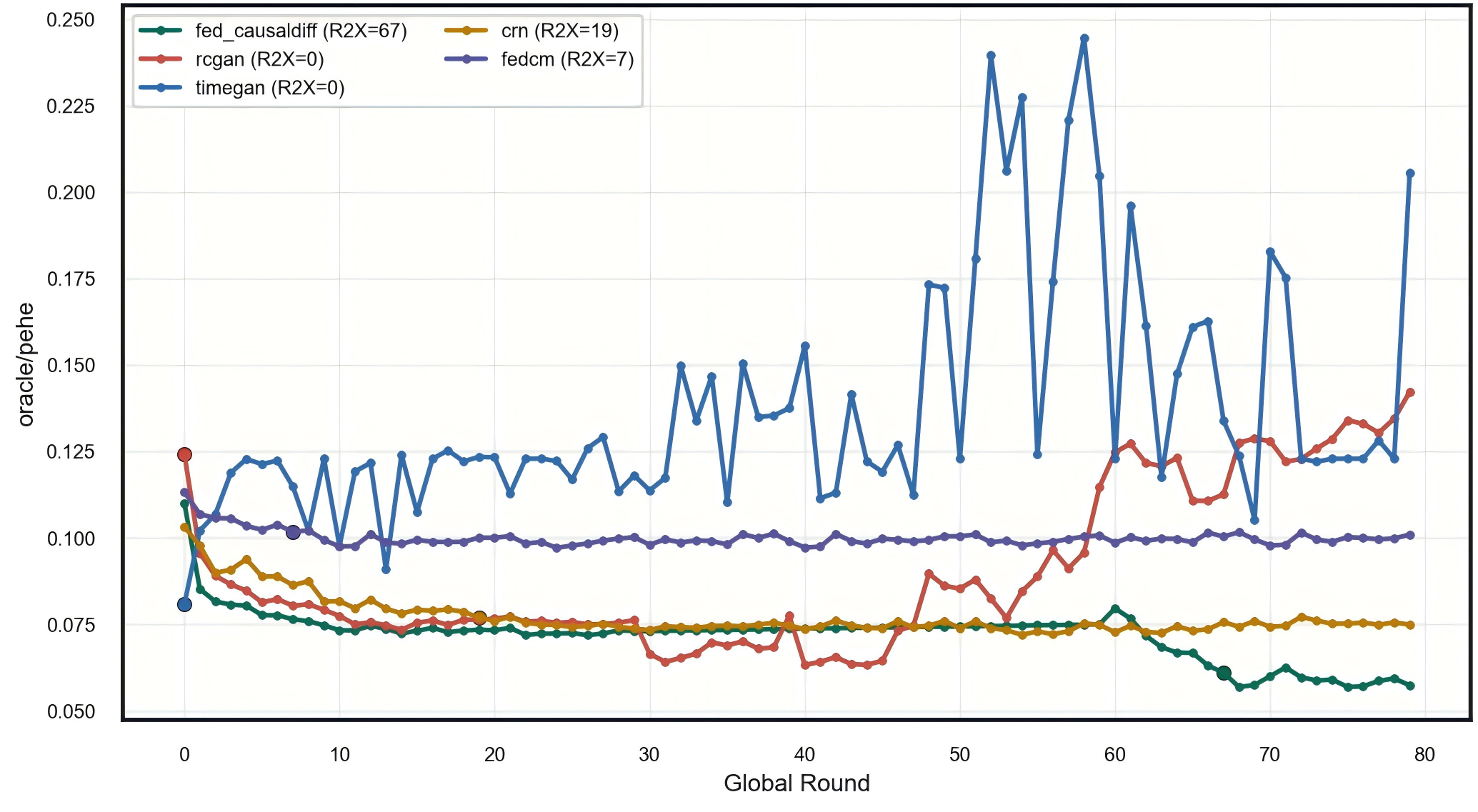}
    \caption{Convergence of interventional accuracy on \textsc{DKT-Synth}. We report the oracle PEHE (lower is better) as a function of the number of global communication rounds. \textsc{Fed-CausalDiff} converges stably and achieves the lowest PEHE, while \textsc{TimeGAN} shows high variance and \textsc{RCGAN} degrades after mid-training.}
    \label{fig:pehe_convergence_dkt}
\end{figure}
Figure~\ref{fig:pehe_convergence_dkt} reports the evolution of the oracle PEHE on \textsc{DKT-Synth} across global communication rounds (lower is better).
\textsc{Fed-CausalDiff} reduces PEHE rapidly in early rounds and continues to improve with stable convergence, eventually reaching the lowest PEHE among all methods.
In contrast, \textsc{TimeGAN} exhibits large oscillations and occasional spikes, indicating unstable interventional learning under federated non-IID training.
\textsc{RCGAN} improves initially but degrades after mid-training, suggesting limited robustness during prolonged federated optimization.
Overall, these trends show that our method achieves better \emph{rounds-to-target} performance, i.e., fewer global rounds are needed to reach a given PEHE threshold.

\paragraph{Communication--performance trade-off.}
\begin{figure*}[t]
    \centering
    \includegraphics[width=0.98\textwidth]{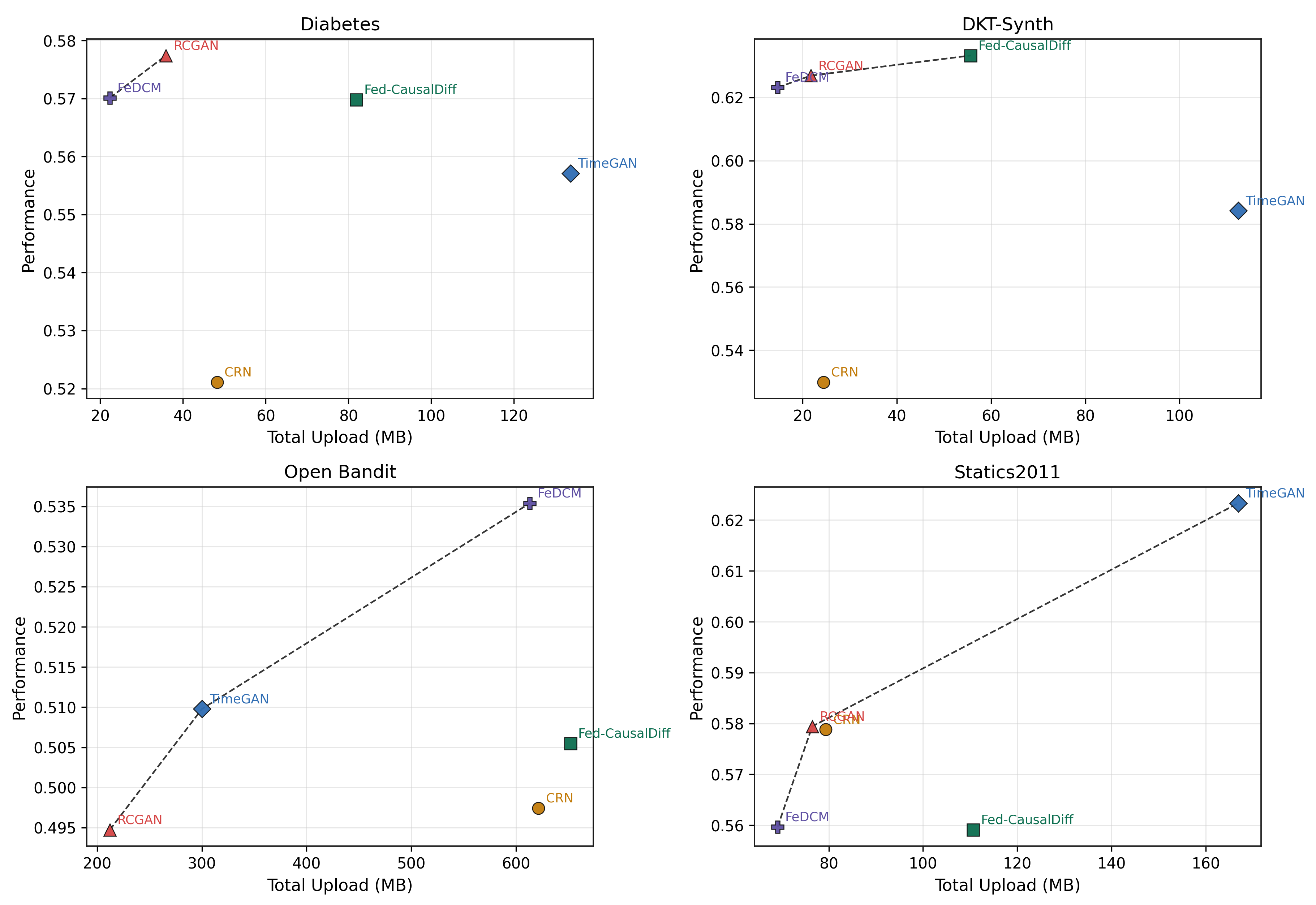}
   \caption{Communication--performance trade-off across datasets. Each point reports a method's downstream AUC performance relative to the total uplink communication (MB) accumulated across all global rounds.}
    \label{fig:comm_perf_tradeoff}
\end{figure*}
Figure~\ref{fig:comm_perf_tradeoff} visualizes downstream performance against the total uplink communication volume (MB).
The trade-off is dataset-dependent: on \textsc{DKT-Synth}, \textsc{Fed-CausalDiff} achieves good performance with a moderate upload budget, whereas \textsc{TimeGAN} attains lower performance despite a substantially higher upload budget.
On \textsc{Diabetes}, \textsc{Fed-CausalDiff} achieves performance comparable to \textsc{FeDCM} while requiring less communication, and uses less upload than \textsc{TimeGAN}.
On \textsc{Open Bandit} and \textsc{Statics2011}, \textsc{Fed-CausalDiff} is not on the Pareto frontier: it requires more (or comparable) upload without improving performance over the best-performing or more communication-efficient baselines.

\section{Discussion}

The results highlight a consistent pattern across metrics. On DKT-Synth, Fed-CausalDiff achieves the lowest errors across all metrics, including PEHE, ATE, and policy value (Table \ref{tab:results_interventional_oracle}). This suggests that its design successfully jointly optimises individual-level counterfactual accuracy (PEHE) alongside population-level effect recovery and long-horizon value, overcoming the trade-offs often seen in other sampleable simulators. In particular, while TimeGAN attains a competitive ATE, it suffers from a substantially larger PEHE and policy-value error, which is consistent with its primary objective of matching observational sequence distributions rather than enforcing interventional/value semantics \cite{yoon2019time,esteban2017real}. The two ablations (\textbf{w/o DSS} and \textbf{w/o $L_{\mathrm{dis}}$}) consistently increase PEHE and substantially enlarge ATE/value errors, further supporting the necessity of our partial-sharing synchronisation and disentanglement objective for calibrated intervention and long-horizon value recovery.

For offline policy value (Fig. \ref{fig:dr-policy-values}), Fed-CausalDiff provides higher mean DR estimates on Diabetes and DKT-Synth, and the Diabetes \enquote{Always} policy returns NaN---an outcome consistent with the practical requirement of sufficient support/overlap for importance-weighting-based estimators. When the logged data provide little or no coverage for actions implied by a target policy, OPE estimates can become unstable or undefined \cite{jiang2016doubly}. On OpenBandit, CRN produces DR values that are markedly larger than those of other methods. Since real-world datasets lack ground-truth returns for counterfactual policies, these magnitudes should be interpreted as estimator outputs rather than verified performance; in practice, the plausibility of policy ranking and the sensitivity to propensities/coverage need domain-side validation \cite{bica2020estimating}.

Table \ref{tab:results_factual_trts_tstr} shows that Fed-CausalDiff remains competitive on TSTR across datasets, while generative baselines (e.g., RCGAN/TimeGAN) often achieve the best purely factual realism scores---consistent with how time-series GANs are designed and commonly evaluated \cite{esteban2017real,yoon2019time}. This supports the intended positioning of Fed-CausalDiff: it prioritises interventional and value-related fidelity while maintaining a practical level of downstream utility, rather than optimising solely for observational realism.

The system efficiency results highlight a dataset-dependent trade-off between communication overhead and causal learning. \textsc{Fed-CausalDiff} demonstrates superior stability and faster convergence in interventional accuracy, effectively minimising the rounds required to reach target PEHE levels compared to the volatile trajectories of \textsc{TimeGAN} and \textsc{RCGAN} (Fig.~\ref{fig:pehe_convergence_dkt}). Regarding communication costs, \textsc{Fed-CausalDiff} proves highly efficient across datasets such as \textsc{DKT-Synth} and \textsc{Diabetes}, achieving top-tier performance with moderate upload budgets. However, its efficiency is not universal. On \textsc{Open Bandit} and \textsc{Statics2011}, it falls short of the Pareto frontier, demanding comparable or greater communication without strictly outperforming lighter baselines (Fig.~\ref{fig:comm_perf_tradeoff}). This indicates that while our method accelerates causal learning and stabilizes convergence, its communication efficiency remains sensitive to the underlying dataset complexity and structural characteristics.

\section{Limitations and Future Work}
A primary limitation of this work is the lack of ground-truth counterfactuals in real-world datasets, aside from DKT-Synth. This necessitates reliance on offline estimators, such as DR or MSM, for intervention and policy value assessment---estimates that must be carefully validated against domain expertise in practical applications. Furthermore, our current experimental setup simulates federated learning on a single machine using FedAvg. It does not yet account for complex system-level factors like differential privacy, asynchronous updates, or client dropouts. Future research will therefore prioritise: (i) developing more robust federated causal evaluation methods with uncertainty quantification; (ii) implementing conservative policy evaluation and coverage constraints for large action spaces; and (iii) exploring personalised or clustered federated learning to enhance fairness and generalisation in non-IID settings.

\section{Conclusion}
Targeting decentralised sequential logs under strict privacy constraints, we introduce Fed-CausalDiff, a federated framework for training interventional sequential causal models. Built upon a unified Dynamic SCM architecture, our approach simultaneously supports factual fitting, $do(\cdot)$ intervention simulation, and closed-loop policy value evaluation. Empirical results demonstrate that Fed-CausalDiff significantly reduces ATE and policy value errors on the DKT-Synth dataset, where counterfactual ground truth is available. Furthermore, on real-world data, it provides more discriminative offline policy value estimates while maintaining competitive TSTR utility. Additionally, system efficiency analyses reveal that while the framework accelerates causal learning with stable convergence, its communication-performance trade-off remains dataset-dependent. Collectively, these findings validate the potential of constructing \enquote{intervenable sequential world models} within federated settings to enable reliable decision evaluation.

\section*{Acknowledgment}
This research is funded by Akademiaavtalen (Equinor ASA) under the ASPIRE (Accelerating Privacy and Data Protection Measures Using Synthetic Data Generation) project.

\bibliographystyle{IEEEtran}
\bibliography{reference}

@inproceedings{mcmahan2017communication,
  title={Communication-efficient learning of deep networks from decentralized data},
  author={McMahan, Brendan and Moore, Eider and Ramage, Daniel and Hampson, Seth and y Arcas, Blaise Aguera},
  booktitle={Artificial intelligence and statistics},
  pages={1273--1282},
  year={2017},
  organization={PMLR}
}

@article{li2020federated,
  title={Federated optimization in heterogeneous networks},
  author={Li, Tian and Sahu, Anit Kumar and Zaheer, Manzil and Sanjabi, Maziar and Talwalkar, Ameet and Smith, Virginia},
  journal={Proceedings of Machine learning and systems},
  volume={2},
  pages={429--450},
  year={2020}
}

@article{wang2020tackling,
  title={Tackling the objective inconsistency problem in heterogeneous federated optimization},
  author={Wang, Jianyu and Liu, Qinghua and Liang, Hao and Joshi, Gauri and Poor, H Vincent},
  journal={Advances in neural information processing systems},
  volume={33},
  pages={7611--7623},
  year={2020}
}

@article{rasouli2020fedgan,
  title={Fedgan: Federated generative adversarial networks for distributed data},
  author={Rasouli, Mohammad and Sun, Tao and Rajagopal, Ram},
  journal={arXiv preprint arXiv:2006.07228},
  year={2020}
}

@article{yoon2019time,
  title={Time-series generative adversarial networks},
  author={Yoon, Jinsung and Jarrett, Daniel and Van der Schaar, Mihaela},
  journal={Advances in neural information processing systems},
  volume={32},
  year={2019}
}

@misc{hernan2010causal,
  title={Causal inference},
  author={Hern{\'a}n, Miguel A and Robins, James M},
  year={2010},
  publisher={CRC Boca Raton, FL}
}

@misc{robins2000marginal,
  title={Marginal structural models and causal inference in epidemiology},
  author={Robins, James M and Hernan, Miguel Angel and Brumback, Babette},
  journal={Epidemiology},
  volume={11},
  number={5},
  pages={550--560},
  year={2000},
  publisher={Lww}
}

@article{lim2018forecasting,
  title={Forecasting treatment responses over time using recurrent marginal structural networks},
  author={Lim, Bryan},
  journal={Advances in neural information processing systems},
  volume={31},
  year={2018}
}

@article{bica2020estimating,
  title={Estimating counterfactual treatment outcomes over time through adversarially balanced representations},
  author={Bica, Ioana and Alaa, Ahmed M and Jordon, James and Van Der Schaar, Mihaela},
  journal={arXiv preprint arXiv:2002.04083},
  year={2020}
}

@article{uehara2022review,
  title={A review of off-policy evaluation in reinforcement learning},
  author={Uehara, Masatoshi and Shi, Chengchun and Kallus, Nathan},
  journal={arXiv preprint arXiv:2212.06355},
  year={2022}
}

@inproceedings{jiang2016doubly,
  title={Doubly robust off-policy value evaluation for reinforcement learning},
  author={Jiang, Nan and Li, Lihong},
  booktitle={International conference on machine learning},
  pages={652--661},
  year={2016},
  organization={PMLR}
}

@article{xiong2023federated,
  title={Federated causal inference in heterogeneous observational data},
  author={Xiong, Ruoxuan and Koenecke, Allison and Powell, Michael and Shen, Zhu and Vogelstein, Joshua T and Athey, Susan},
  journal={Statistics in Medicine},
  volume={42},
  number={24},
  pages={4418--4439},
  year={2023},
  publisher={Wiley Online Library}
}

@article{khellaf2024federated,
  title={Federated causal inference: Multi-study ate estimation beyond meta-analysis},
  author={Khellaf, R{\'e}mi and Bellet, Aur{\'e}lien and Josse, Julie},
  journal={arXiv preprint arXiv:2410.16870},
  year={2024}
}

@article{khellaf2025federated,
  title={Federated Causal Inference from Multi-Site Observational Data via Propensity Score Aggregation},
  author={Khellaf, R{\'e}mi and Bellet, Aur{\'e}lien and Josse, Julie},
  journal={arXiv preprint arXiv:2505.17961},
  year={2025}
}

@article{gao2021federated,
  title={Federated causal discovery},
  author={Gao, Erdun and Chen, Junjia and Shen, Li and Liu, Tongliang and Gong, Mingming and Bondell, Howard},
  journal={OpenReview},
  note={ICLR 2022 withdrawn submission},
  year={2021}
}

@inproceedings{wang2023towards,
  title={Towards practical federated causal structure learning},
  author={Wang, Zhaoyu and Ma, Pingchuan and Wang, Shuai},
  booktitle={Joint European Conference on Machine Learning and Knowledge Discovery in Databases},
  pages={351--367},
  year={2023},
  organization={Springer}
}

@article{pearl2012causal,
  title={The causal foundations of structural equation modeling},
  author={Pearl, Judea},
  journal={Handbook of structural equation modeling},
  pages={68--91},
  year={2012}
}

@book{pearl2009causality,
  title={Causality},
  author={Pearl, Judea},
  year={2009},
  publisher={Cambridge university press}
}

@inproceedings{joachims2018deep,
  title={Deep learning with logged bandit feedback},
  author={Joachims, Thorsten and Swaminathan, Adith and De Rijke, Maarten},
  booktitle={International Conference on Learning Representations},
  year={2018}
}

@incollection{de2020principles,
  title={Principles relating to processing of personal data},
  author={De Terwangne, Cecile},
  booktitle={The EU general data protection (GDPR): a commentary},
  pages={309--320},
  year={2020},
  publisher={Oxford University Press}
}

@article{kairouz2021advances,
  title={Advances and open problems in federated learning},
  author={Kairouz, Peter and McMahan, H Brendan and Avent, Brendan and Bellet, Aur{\'e}lien and Bennis, Mehdi and Bhagoji, Arjun Nitin and Bonawitz, Kallista and Charles, Zachary and Cormode, Graham and Cummings, Rachel and others},
  journal={Foundations and trends{\textregistered} in machine learning},
  volume={14},
  number={1--2},
  pages={1--210},
  year={2021},
  publisher={Now Publishers, Inc.}
}

@article{gao2024causal,
  title={Causal inference in recommender systems: A survey and future directions},
  author={Gao, Chen and Zheng, Yu and Wang, Wenjie and Feng, Fuli and He, Xiangnan and Li, Yong},
  journal={ACM Transactions on Information Systems},
  volume={42},
  number={4},
  pages={1--32},
  year={2024},
  publisher={ACM New York, NY}
}

@article{shinozaki2020understanding,
  title={Understanding marginal structural models for time-varying exposures: pitfalls and tips},
  author={Shinozaki, Tomohiro and Suzuki, Etsuji},
  journal={Journal of epidemiology},
  volume={30},
  number={9},
  pages={377--389},
  year={2020},
  publisher={Japan Epidemiological Association}
}

@article{xie2018off,
  title={Off-policy evaluation and learning from logged bandit feedback: Error reduction via surrogate policy},
  author={Xie, Yuan and Liu, Boyi and Liu, Qiang and Wang, Zhaoran and Zhou, Yuan and Peng, Jian},
  journal={arXiv preprint arXiv:1808.00232},
  year={2018}
}

@inproceedings{ghosh2020context,
  title={Context-aware attentive knowledge tracing},
  author={Ghosh, Aritra and Heffernan, Neil and Lan, Andrew S},
  booktitle={Proceedings of the 26th ACM SIGKDD international conference on knowledge discovery \& data mining},
  pages={2330--2339},
  year={2020}
}

@article{strack2014impact,
  title={Impact of HbA1c measurement on hospital readmission rates: analysis of 70,000 clinical database patient records},
  author={Strack, Beata and DeShazo, Jonathan P and Gennings, Chris and Olmo, Juan L and Ventura, Sebastian and Cios, Krzysztof J and Clore, John N},
  journal={BioMed research international},
  volume={2014},
  number={1},
  pages={781670},
  year={2014},
  publisher={Wiley Online Library}
}

@article{saito2020open,
  title={Open bandit dataset and pipeline: Towards realistic and reproducible off-policy evaluation},
  author={Saito, Yuta and Aihara, Shunsuke and Matsutani, Megumi and Narita, Yusuke},
  journal={arXiv preprint arXiv:2008.07146},
  year={2020}
}

@article{esteban2017real,
  title={Real-valued (medical) time series generation with recurrent conditional gans},
  author={Esteban, Crist{\'o}bal and Hyland, Stephanie L and R{\"a}tsch, Gunnar},
  journal={arXiv preprint arXiv:1706.02633},
  year={2017}
}

@article{piech2015deep,
  title={Deep knowledge tracing},
  author={Piech, Chris and Bassen, Jonathan and Huang, Jonathan and Ganguli, Surya and Sahami, Mehran and Guibas, Leonidas J and Sohl-Dickstein, Jascha},
  journal={Advances in neural information processing systems},
  volume={28},
  year={2015}
}

@inproceedings{shalit2017estimating,
  title={Estimating individual treatment effect: generalization bounds and algorithms},
  author={Shalit, Uri and Johansson, Fredrik D and Sontag, David},
  booktitle={International conference on machine learning},
  pages={3076--3085},
  year={2017},
  organization={PMLR}
}

@inproceedings{collins2021exploiting,
  title={Exploiting shared representations for personalized federated learning},
  author={Collins, Liam and Hassani, Hamed and Mokhtari, Aryan and Shakkottai, Sanjay},
  booktitle={International conference on machine learning},
  pages={2089--2099},
  year={2021},
  organization={PMLR}
}

@article{amalan2022multi,
  title={Multi-flgans: multi-distributed adversarial networks for non-IID distribution},
  author={Amalan, Akash and Wang, Rui and Qiao, Yanqi and Panaousis, Emmanouil and Liang, Kaitai},
  journal={arXiv preprint arXiv:2206.12178},
  year={2022}
}

@article{peng2025federated,
  title={Federated Learning for Diffusion Models},
  author={Peng, Zihao and Wang, Xijun and Chen, Shengbo and Rao, Hong and Shen, Cong and Jiang, Jinpeng},
  journal={IEEE Transactions on Cognitive Communications and Networking},
  year={2025},
  publisher={IEEE}
}

@inproceedings{rahman2025fedcm,
  title={FeDCM: Federated Learning of Deep Causal Generative Models},
  author={Rahman, Md Musfiqur and Kocaoglu, Murat},
  booktitle={The 41st Conference on Uncertainty in Artificial Intelligence},
  year={2025}
}

\end{document}